# Thoughts on Architecture

Paul S. Rosenbloom

Department of Computer Science
University of Southern California, Los Angeles, CA USA
`rosenbloom@usc.edu`

**Abstract.** The term *architecture* has evolved considerably from its original Greek roots and its application to buildings and computers to its more recent manifestation for minds. This article considers lessons from this history, in terms of a set of relevant distinctions introduced at each of these stages and a definition of architecture that spans all three, and a reconsideration of three key issues from cognitive architectures for architectures in general and cognitive architectures more particularly.

## 1 Introduction

Architectures are central to many attempts to capture key aspects of what is necessary for general intelligence [1-4], whether as models of natural (human) intelligence or as artificial systems that embody human-level intelligence or beyond. This notion is most familiar in cognitive science, where a cognitive architecture is intended to embody a theory of the human mind. However, it can also be found in both artificial intelligence (AI) and artificial general intelligence (AGI) – although sometimes under other names, such as AI, AGI, and agent architectures – where it amounts to a fixed framework that supports the construction of (generally) intelligent systems. For simplicity, the term cognitive architecture is used here as the generic across all of these variants.

One definition of a cognitive architecture is as a *hypothesis concerning the fixed structures and processes that together yield a mind, whether natural or artificial* [5]. Explicit in this definition is that a cognitive architecture should be *fixed* and that it may concern *natural or artificial minds*. Implicit in it is that a cognitive architecture should be *theoretical* ("hypothesis concerning") and *computational* ("structures and processes"). However, computer architectures, the direct inspiration for cognitive architectures, although fixed and computational, are not obviously theoretical in the sense of being *about* something else. Reaching further back in the inspirational chain, building architectures, although fixed, are neither computational nor theoretical.

In his classic work on *Unified Theories of Cognition*, Newell [6] defined an architecture as a *symbol system* [7]. This shares the concern with fixity and the notion that an architecture in general need not be theoretical, but then narrows the definition from there down to a form of universal computation. It thus rules out building architectures and any other noncomputational structures – such as noncomputational scientific theories – or even many special purpose computational devices, while still admitting typical computer architectures and symbolic cognitive architectures. By having identified minds with symbol systems [7] – an idea that remains controversial to date – Newell had no need to add that cognitive architectures are about minds, whereas under the version in [5] they are explicitly concerned with understanding and/or building minds.

The remainder of this article begins with an attempt to clarify the notion of an architecture as it has developed from building architecture, through computer architecture, to cognitive architecture (Section 2). A key part of this is identifying relevant *distinctions* that the kinds of architectures in this sequence have introduced. Analyzing this history, and the distinctions introduced across it, a broad definition of an architecture emerges that spans the entire sequence (Section 3). What then follows are discussions of three issues of interest that specifically concern cognitive architecture: multilayered architectures (Section 4), theoretical computational architectures (Section 5), and architectural exploration (Section 6). The ultimate intent is to see what new insight this all can yield for architecture in general and for cognitive architecture more particularly.

## 2   Historical Context[1]

The word *architecture* stems from the Greek ἀρχιτέκτων (*arkhitekton*), where it referred to a master builder or director of works [8-9]. In modern usage it includes "both the process and the product of planning, designing, and constructing buildings or other structures" [9], with its practice oriented toward both practical and expressive requirements, and thus serving both utilitarian and aesthetic ends [10]. Distinctions that arise in such architectures that resonate in later forms of architecture include:

1. The *fixed* nature of the architecture versus the *variable* nature of its contents. In French, this is succinctly *immobiliere* (real estate) versus *mobiliere* (furniture).
2. *Design* versus *implementation*. A design is part of the process that specifies what is to be implemented, whereas an implementation is the heart of the resulting product.
3. (*Function* versus *structure*) versus *form*. Function concerns how the product is used whereas structure concerns how this function is provided. Both relate to utility whereas form in this context concerns the aesthetics of how these are presented.
4. *Simple* versus *complex*. Although this may impact utility, the greater concern is with form/aesthetics, such as Scandinavian or Japanese versus Victorian or rococo.

*Computer architecture* [11] is a more recent conceptualization inspired by the notion of building architecture. It induces a partitioning of a computational system – i.e., a system that transforms information [12] – into a fixed architecture versus variable programs and data. The design is a description whereas the implementation actively computes. The earliest computers each had their own *instruction set* – a specification of the instructions the computer could execute – that was tied to the hardware they came with.

---

[1]   *Truth in advertising:* I am an expert on cognitive architecture, but nothing more than a student of computer architecture and an interested outsider on building architecture.

However, it was not long before this was separated from the implementation, with the architecturally defined instruction set specifying what was to be implemented, and multiple hardware implementations being developed to span sizes and generations of computers. A general language – Instruction Set Processor (ISP) – was even developed to enable specifying instruction sets across the field of computer architecture [13].

The function of a computer architecture is to support computation via an instruction set. The structure comprises the hardware (and possibly firmware) components that yield this plus their organization. Form is often deprecated, although some manufacturers do take it seriously. There is an analog of simple versus complex, in the debate between reduced versus complex instruction set computers – i.e., RISC versus CISC – although the focus here is more on utility than aesthetics.

A computer architecture induces one additional distinction of note:

5. *Transformer* versus *container*. A transformational architecture transforms its variable content rather than simply containing it.

A computational architecture transforms information, with a computer architecture further specializing this to be based on hardware (and possibly firmware) structures.

The notion of a *cognitive architecture* is even newer, being a form of computational architecture that originated in analogy to computer architecture, with Allen Newell having played formative roles in both areas [13-14]. Its function is to provide a (fixed) *mind* that can transform (variable) *knowledge and skills*. Its structure comprises the mechanisms and their organization that together yield a mind. The question of form arises primarily in terms of the simple versus complex distinction. A simple, RISC-like cognitive architecture focuses on achieving intelligence from the interactions among a small number of very general mechanisms (see, e.g., [5, 15]), whereas a complex, CISC-like architecture includes a wider range of more complex mechanisms (see, e.g., [16-17]). This distinction may be thought of as between a physics (i.e., beauty) mindset – although even there it can be controversial [18] – versus a biological (i.e., evolution-as-a-tinkerer) one, with the computer science (i.e., modular) perspective possibly sitting in between. Both utilitarian and aesthetic aspects are implicated here.

The introduction of cognitive architecture also induces an additional distinction:

6. *Theoretical* versus *atheoretical*. This amounts to whether or not an architecture is *about* something else. Architectures in cognitive science tend to be theoretical – being about human minds – whereas A(G)I architectures (along with building and computer architectures) are not directly so.

A theoretical architecture – or *theory* – with or without accompanying variations, represents – or "stands in for" – key aspects of a phenomenon or domain of interest. Architectures represent complex phenomena when used in *understanding* it [12], as is typical in the sciences. Transformational architectures may also be used to generate (or *shape* [12]) complex phenomena, as is typical in engineering and other professions [12]. Such architectures may also represent something but need not do so.

A computer architecture, for example, is central to engineering all kinds of computational systems, with its foremost purpose being to aid in the development of such systems rather than to represent or to help in understanding anything. The architecture in such a case could conceivably be considered as a model of computation, as with a Turing machine [19], but this is deep in the background in most applications. Similarly, a cognitive architecture may be theoretical in a strong sense if intended, for example, to explicitly represent a human mind or it may be theoretical only in a weaker sense, in implicitly embodying a hypothesis about mind in the abstract. The definition in the introduction is broad enough to span both senses.

An *inspired* architecture, such as a biologically inspired cognitive architegure (BICA) [20], sits at hybrid position with respect to theoretical versus atheoretical via a particular combination of understanding and shaping. At the top level, the goal is atheoretical, to support the development of useful systems with no concern as to what might be represented in the process. However, understanding of an existing system becomes an instrumental subgoal in guiding the design of the atheoretical architecture. In some cases the existence of this subsidiary understanding is considered purely heuristic, and fine to ignore or dismiss when convenient, while in others it is considered a virtue to be extolled.

## 3    Defining Architecture

From these three overall classes of architectures – building, computer, and cognitive – a slimmed-down definition of an architecture can be identified that centers on a *fixed framework* that enables and delimits a *space of variations*. This focuses on Distinction 1, with the other five helping to scope the space of architectural types while not themselves being definitional. Simon [21] discussed the importance of identifying *invariants* because of "their power to strip away the complexity and diversity of a whole range of phenomena and to reveal the simplicity and order underneath." In this sense, an architecture is a set of invariants; however, it also typically goes beyond this to consider the interactions among them and what a complete set of them might be.

Together, an architecture and a set of variations yields a *system*, which can succinctly be semi-formalized via the equation *system = architecture + variations*. This harks back to earlier syntactically similar equations, such as *algorithms + data structures = programs* [22] and *algorithm = logic + control* [23]; however, each of these equations makes a distinct point. Still, in the spirit of the second equation, the familiar equation *system(s) = program(s) + data* can be seen as a particular specialization of the equation proposed here to computational systems, where the program is the fixed, architectural component and the data yields the variations.

More broadly, the systems of interest here may be conceptual, mathematical, computational, physical, etc. An *application* is then simply a system that yields value. In general, different kinds of architectures enable different forms of variations (and thus yield different types of systems and applications). As we have just seen, computer architectures enable programs and data while cognitive architectures enable skills and knowledge. In addition, modular architectures enable module definitions and API (i.e., application programming interface) specifications; hierarchical and graphical architectures enable node and link definitions; tables and maps enable entries at appropriate locations; and buildings enable furniture, appliances, and ornaments. In tightly specified theories or models, in which nearly the entire system is architecture, the variations may simply be parameter values.

A classic slogan in AI, although it seems to have originated hundreds of years ago with Francis Bacon as *scientia potentia est*, is that *knowledge is power* [24]. We could also add that *applications are payoff*. In architectural terms, knowledge is a form of variation enabled by cognitive architectures, whereas applications are complete systems that result from combining appropriate architectures and variations. In analogy, we can now go a step further to say that *architecture is essence*.

## 4 Multilayered Architectures

Computational architectures – i.e., architectures that transform information – and theoretical architectures are both amenable to being parts of layered stories.[2] Consider first computational architectures. A computer architecture provides a fixed foundation that supports leveraging programs to transform data. However, as anticipated with the familiar equation introduced in the previous section, at a second level programs may themselves yield computational architectures, particularly when fixed over some span of interest, with their data then providing the only source of variation. This data variation may in turn be very flexible – as for example in a cognitive architecture that supports a broad range of skills and knowledge – or it may be severely limited.

In the former case, this process of implementing computational architectures within the variations that are enabled by other computational architectures can conceivably proceed to arbitrary levels. For example, the Sigma cognitive architecture [5] comprises two such layers – a *graphical* architecture and a *cognitive* architecture – with the former implementing the latter. Some of the capabilities included in the Common Model of Cognition [3] – such as a particular form of declarative memory – are then implemented in a third layer, with the aid of skills and knowledge represented within Sigma's cognitive architecture.

Similarly, a theoretical architecture can be a theory itself or just one part of a more elaborate theory that also includes appropriate choices among the variations the architecture enables. Kivunja [25], for example, discusses the notion of a theoretical framework as the accepted wisdom from experts that serves as the background for a graduate student's own research. The former might perhaps also be considered a Kuhnian paradigm [26]. This effectively partitions a student's contributed theory into a fixed, architectural framework – or paradigm – plus the student's own contributed variations. A Lakatosian research programme [27] also partitions in this manner – "For Lakatos an individual theory within a research programme typically consists of two components: the (more or less) irrefutable hard core plus a set of auxiliary hypotheses" [28].

A cognitive architecture can be considered as a theory of the fixed structures and processes found in a human mind, or of what is necessary and/or sufficient to yield an artificial mind. When combined with skills and knowledge, it can also serve as a more fleshed out model, for example, of human behavior in a particular task. While cognitive architectures are often both computational and theoretical, only the more fleshed out models can execute to yield behavioral data for comparison with human data, or applications that have value beyond their ability to model humans.

---

[2] The same may also be true beyond computational architectures to all transformational architectures, but the focus here is narrower.

## 5   Theoretical Computational Architectures

For theoretical computational architectures, there is a serious question of how their design, implementation and theory are related; and, also, which of these is properly an/the architecture. In general, a design is a description, and an implementation is a realization of this description. For buildings, both are architecture, whereas for computers only the design is, with the implementation being instead a computer. For cognition the implementation is typically the architecture, as there is rarely an explicit design. However, for a theoretical cognitive architecture this yields a conundrum, as there often is a vague notion of what the architecture is beyond any particular implementation that enables researchers to refer to multiple implementations, even in different underlying languages, as being the same architecture.

Starting back at the beginning, in the simplest case there is just a theory about a body of phenomena (Fig. 1). It might be described informally in text or more formally as mathematical equations. For a computational theory, there is typically an implementation, of which the theory itself is only a subpart. The remainder is "implementation details" necessary to make the theory executable but not part of the theory itself (Fig. 2). There is typically no separate design specification in such cases, although AIXI [15] could be considered as such, and there were several attempts to specify Soar more formally [29-30]. Still, ideally there would always be a separate design that specifies in a more comprehensible yet abstract fashion what is to be in the implementation (Fig. 3), as is common with buildings and computers. This design would have its own theoretical subpart – corresponding to the theory in Fig. 1 – which may be more easily demarcated than is possible within the implementation. Without this, the boundary between what is part of the theory versus an implementation detail remains fuzzy [30].

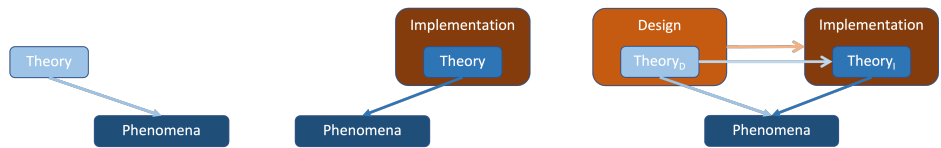

**Fig. 1.** Simple theory.     **Fig. 2.** Typical computational theory.     **Fig. 3.** Ideal computational theory.

Where is the architecture in all of this? According to the definition in Section 3, all four non-phenomenal components in Fig. 3 are theoretical architectures. They may even share the same name in practice, although Fig. 3 can potentially be leveraged to distinguish the distinct roles they each fill with respect to a common name. It should be noted though that the atheoretical aspects of the two larger components maintain an ambiguous state of being fixed with respect to their enabled variations but malleable in being adjustable as necessary without affecting the core theory, much like Lakatosian auxiliary hypotheses. The two design architectures in turn can be doubly theoretical, in being both about the phenomena and the implementation.

## 6 Architectural Exploration

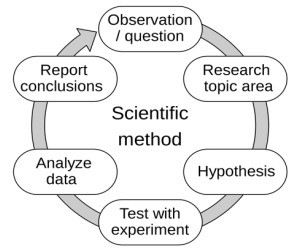

Fig. 4. The scientific method [31].

Fig. 4 shows one way of conceptualizing the traditional *scientific method*. As parts of a method diagram, these boxes are an odd mix of nouns and verbs, although they do include the standard pieces in one form or another. Still, when considering my own architectural methodology over the past 40+ years, I end up with something more like Fig. 5, which labels nouns as nodes while relegating verbs to arrows. According to this view, there is a constant back and forth between an architecture and its researchers, with the former yielding new insight for the latter and in return being modified by them to capture this insight. The phenomena of interest are themselves understood (theoretical) and/or shaped (atheoretical) by the architecture, with this interaction providing input for both directions of the architecture-researchers interaction. There is also a direct link between researchers and phenomena that reflects both architecture-free observation and direct exploration of the phenomena.

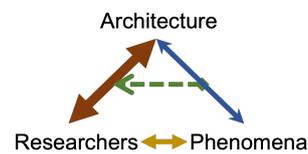

Fig. 5. Architectural exploration.

*Exploration* – i.e., searching for new phenomena to understand – has no explicit role in Fig. 4, although it may be implicit within the *Observation/question* box. It is often considered pre-scientific, but the scientific enterprise would be terribly impoverished without it. Early natural scientists fanned out across the globe to deliberately seek out and make sense of novel plants, animals, and other natural phenomena. To this day, disciplines such as paleontology, astronomy and particle physics intentionally explore the natural universe. They may also expend considerable effort developing new instruments that sense aspects of the world/universe not previously perceptible to us. There may be hypotheses and predictions that are intended from the beginning to be tested via such instruments, but part of the anticipation and excitement in any such enterprise must always be the possibility of uncovering what was not expected.

In Fig. 5, the researchers-phenomena arrow represents classical exploration as a form of active observation, with the leftward direction reflecting observation and the rightward direction manipulation of what is to be observed (with controlled experimentation at the extreme). The architecture-researchers arrow then represents architectural exploration. The downward direction represents the development of insight based on the architecture and its interactions with the phenomena of interest (with direct observation entering from the side). It also reflects the architecture's role as both an instrument and a guide. The upward direction represents architecture modification based on the existing architecture and its interactions with the phenomena of interest. This reflects the architecture's role as both a hypothesis and a domain of exploration of its own.

The cognitive architectures I have developed over the years have partially been intended as expressions of, and means for testing, hypotheses. Often these hypotheses start out rather vague,[3] such as some notion of the potential benefits deriving from the combination of rules and neural-like activation in Xaps [33]; rules, problem solving and learning-by-chunking in Soar [34]; and Soar-like cognitive architectures and probabilistic graphical models in Sigma [5]. Sometimes this develops into a somewhat more precise – and even at times grander – hypothesis, such as that chunking provides a general learning mechanism [35] or that Soar can support a *unified theory of cognition* [6]. More often, crisper small hypotheses are spun off, such as that chunking in Soar could be used not just to speed up performance but also to learn new things [36], or that Sigma could straightforwardly be extended to incorporate neural networks (NNs) [37].

Still, most of the actual work is exploratory, using the architecture to guide the exploration of cognition and to interpret the results of these explorations; as well as searching the space of cognitive architectures itself. Controlled experiments are relevant to only a fraction of the criteria used to evaluate such work, with the more complete criteria including: (1) whether an architecture functions as intended; (2) how broadly the architecture models and/or produces the phenomena of interest; (3) how simple and elegant the architecture is; and (4) how much insight the architecture inspires and captures, whether into anticipated implications or wholly unanticipated ones.

Despite numerous attempts by the field to develop better approaches to evaluating work on cognitive architectures, much of the evaluation activity around these four criteria necessarily continues to center around simply building new architectural components and exploring their interactions. Controlled experiments, in contrast, tend to focus on refining what can be said about criterion (1), in terms of how well the architecture models and/or produces phenomena of interest, and a bit of criterion (2). Such experiments clearly play a role, but not particularly a dominant one. If one is problem focused, rather than methodology focused, it is critical to use whatever the best methods available are for the problem of interest, rather than limiting oneself to problems for which the strongest methods – in terms of the veracity they guarantee – are applicable; and exploration is still the best methodology for much work in cognitive architectures.

Exploration was the name of the game in the early days of AI, when there were many more unexplored areas than researchers to investigate them and it was better to move on to new topics once sufficient low-hanging fruit had been harvested than to work over existing topics very carefully first. Now, however, exploration has fallen into relative disrepute as a method of investigating topics in AI. With a broader coverage of the space of topics now under our belts, and general progress in the application of proofs and controlled experiments to specific AI problems, these methodologies have become the coin of the realm instead. Strong methods once they get a foothold in a field tend to push out weaker ones even in areas where the stronger ones are not applicable [12].

Exploration has continued to remain a major methodology in AGI, presumably enabled by separating the field out from AI as a distinct discipline. However, in doing so, it must also be sure to avoid becoming a wayside by either ignoring what is happening in the mainstream or by clinging to weaker methods if/when stronger ones do come along that apply to the problem(s) of producing general intelligence.

---

[3] "I do not pretend to start with precise questions. I do not think you can start with anything precise. You have to achieve such precision as you can, as you go along." [32].

## 7       Conclusion

The crux of the idea of an architecture is to differentiate the fixed from the variable aspects of a system. This distinction, which in honor of Simon might be denoted simply as *(in)variant*, or even as *(im)mobiliere*, can then be elaborated on in various ways, with the (non)computational and (a)theoretical distinctions being of particular relevance to cognitive architectures. Additional issues that have been revisited here concern architectural layering and the nature of theoretical computational architectures. The question of how architectures should be investigated, with a specific focus on exploration, has then capped off these thoughts. Future work might occur in a variety of directions, including incorporating into this analysis other forms of architecture, such as software architecture [38-39], that are relevant to architecture in general, and that should in principle be relevant to cognitive architecture, but which did not have a foundational impact on the development of this latter notion.